\documentclass[letterpaper]{article} 
\usepackage{aaai25}  
\usepackage{times}  
\usepackage{helvet}  
\usepackage{courier}  
\usepackage[hyphens]{url}  
\usepackage{graphicx} 
\usepackage{natbib}  
\usepackage{caption} 
\usepackage{algorithm}
\usepackage{algorithmic}
\usepackage{amsmath,amssymb,amsfonts}
\usepackage{multirow}

 \usepackage{booktabs}

\usepackage{newfloat}
\usepackage{listings}
\DeclareCaptionStyle{ruled}{labelfont=normalfont,labelsep=colon,strut=off} 
\floatstyle{ruled}
\newfloat{listing}{tb}{lst}{}
\floatname{listing}{Listing}
\nocopyright

\title{
    \begin{minipage}[c]{0.12\textwidth}
        \includegraphics[width=\textwidth]{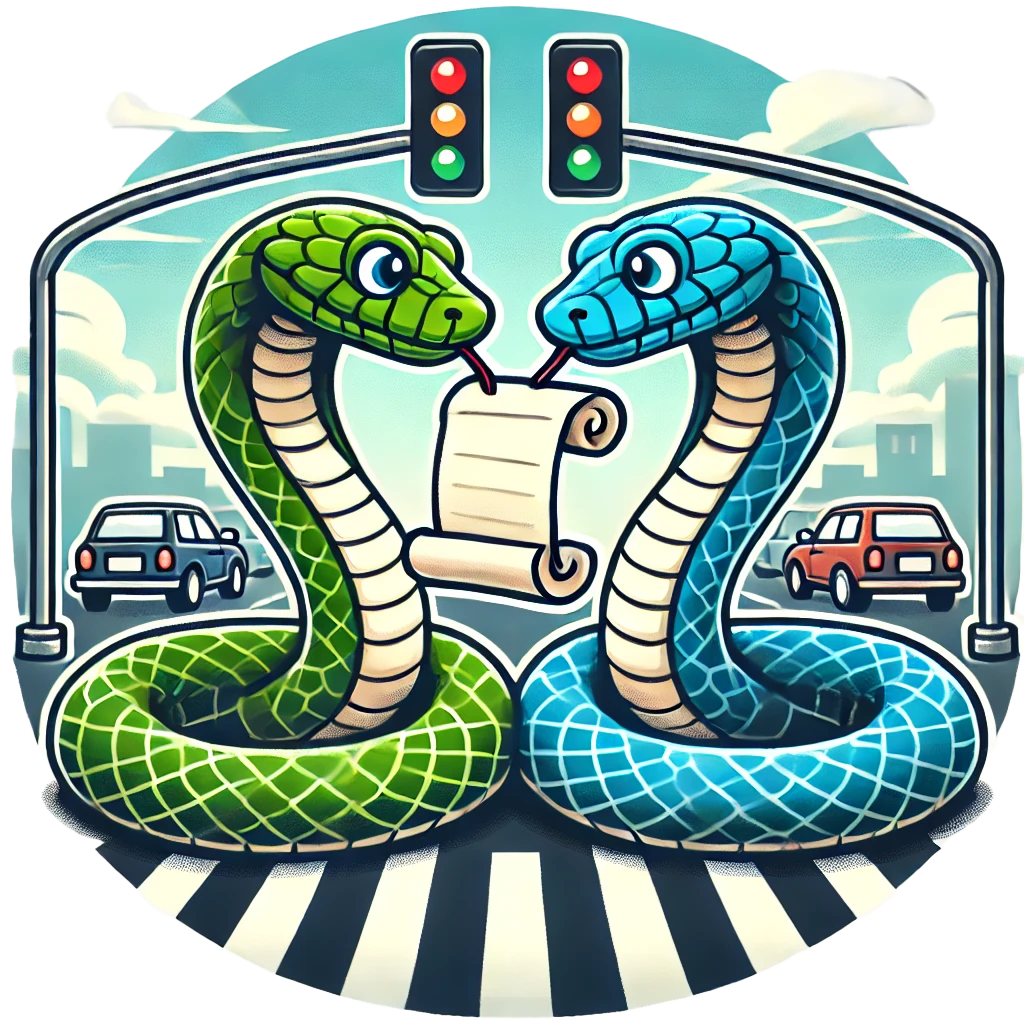}
    \end{minipage}%
    \begin{minipage}[c]{0.88\textwidth}
        \centering
	CollaMamba: Efficient Collaborative Perception with Cross-Agent Spatial-Temporal State Space Model
    \end{minipage}
}

\author{
    Yang Li, Quan Yuan, Guiyang Luo, Xiaoyuan Fu, Xuanhan Zhu, Yujia Yang, Rui Pan, Jinglin Li
}
\affiliations{

    State Key Laboratory of Networking and Switching Technology, Beijing University of Posts and Telecommunications \\
    \{leeyang866, yuanquan, luoguiyang, fuxiaoyuan, xuanhan.zhu, yangyj2022, panrui805, jlli\}@bupt.edu.cn

}

\usepackage{bibentry}

\begin{document}

\maketitle

\begin{abstract}
By sharing complementary perceptual information, multi-agent collaborative perception fosters a deeper understanding of the environment. 
Recent studies on collaborative perception mostly utilize CNNs or Transformers to learn feature representation and fusion in the spatial dimension, which struggle to handle long-range spatial-temporal features under limited computing and communication resources. 
Holistically modeling the dependencies over extensive spatial areas and extended temporal frames is crucial to enhancing feature quality. 
To this end, we propose a resource efficient cross-agent spatial-temporal collaborative state space model (SSM), named CollaMamba. 
Initially, we construct a foundational backbone network based on spatial SSM. This backbone adeptly captures positional causal dependencies from both single-agent and cross-agent views, yielding compact and comprehensive intermediate features while maintaining linear complexity. 
Furthermore, we devise a history-aware feature boosting module based on temporal SSM, extracting contextual cues from extended historical frames to refine vague features while preserving low overhead. 
Extensive experiments across several datasets demonstrate that CollaMamba outperforms state-of-the-art methods, achieving higher model accuracy while reducing computational and communication overhead by up to $71.9\%$ and $1/64$, respectively.
This work pioneers the exploration of the Mamba’s potential in collaborative perception. The source code will be made available.
\end{abstract}

%

\section{Introduction}
In field such as autonomous driving and intelligent robotics, accurate and comprehensive perception of complex environments by agents is crucial for ensuring safety, reliability, and efficiency \cite{chenEndtoendAutonomousDriving2023}. Single-agent perception, however, is constrained by a limited field of view, leading to reduced accuracy. Multi-agent collaborative perception offers a promising solution to these dilemmas by facilitating the exchange of complementary perceptual information among agents, thereby enhancing their individual perception capabilities \cite{yazganSurveyIntermediateFusion2024}.
Recent methods based on intermediate feature fusion have gained considerable attention in collaborative perception \cite{chenMilestonesAutonomousDriving2023, gaoSurveyCollaborativePerception2024}. Some of the approaches are designed to strike a balance between perception performance and communication bandwidth, with notable examples including distillation-based fusion models \cite{liLearningDistilledCollaboration2021a}, attention-based and transformer-based approaches like V2X-ViT \cite{xuV2XViTVehicletoEverythingCooperative2022a}, all of which have demonstrated strong performance. 

\textit{Significant challenges and opportunities for improvement persist in cross-agent collaborative perception, especially in effectively and efficiently modeling holistic long-range spatial-temporal dependencies of the intermediate features.}
First, for single agent, most existing methods rely on CNNs or ResNets to encode observations into 2D intermediate feature maps \cite{hanCollaborativePerceptionAutonomous2023}. However, the limited receptive fields of convolution operations often fail to capture long-range spatial dependencies of the views \cite{zhuModelingLongrangeDependencies2023}. Effectively capturing these intrinsic spatial relationships and causal dependencies could further compress shared features, yielding more compact features and reducing resource overhead.
In addition, for cross-agent collaborative fusion, while multi-scale attention mechanisms and transformers facilitate cross-agent spatial modeling \cite{xuCoBEVTCooperativeBird2022}, they introduce high computational complexity, with computational resource demands increasing quadratically as the input observation dimensions and the number of agents grow.
Moreover, most methods rely on static predictions from the current frame, neglecting the complementary information and motion cues available over longer temporal scales. These cues are highly reliable and play a crucial role in boosting and refining the intermediate features at the current moment. Even with the fusion of one or several frames of historical information, it remains inadequate to fully capture the essential contextual details of the target at the current moment for effective assistance. Even though long-term temporal context is crucial for refining ambiguous features, processing extended spatiotemporal sequences introduces challenges related to computational overhead and latency \cite{liuSelect2ColLeveragingSpatialTemporal2023}.


Recently, state space model (SSM), particularly with the Mamba architecture as an advanced iteration, have garnered significant attention for their ability for long-range sequential modeling. Building on these insights, we propose a unified, resource-efficient cross-agent spatial-temporal collaborative perception architecture based on SSM, named CollaMamba. The overall structure of CollaMamba is illustrated in Figure \ref{fig-overview}. CollaMamba primarily consists of a cross-agent spatial collaboration backbone network, along with two plug-and-play modules: single-agent history-aware feature boosting and cross-agent collaborative prediction.

To address the insufficient extraction of long-range spatial dependencies and the associated high resource overheads, we design the cross-agent spatial collaboration backbone network, which includes a Mamba-based encoder, decoder, and a cross-agent fusion module. The Mamba encoder effectively captures single-agent long-range spatial dependencies, producing compact sequence-form intermediate features that accurately model positional causal relationships, thereby ensuring high feature quality and communication efficiency when sharing these features. 
The cross-agent fusion module precisely captures feature correlations and long-range spatial dependencies across multiple agents and global scene while maintaining robustness to position and latency discrepancies. This fusion approach preserves linear complexity, enabling it to efficiently accommodate a greater number of neighbor agents while keeping computational overhead low.

Additionally, we focus on the temporal dimension and exploit holistic long-range spatial-temporal dependencies to enhance intermediate features. For single-agent tasks, single-agent history-aware feature boosting module leverages long-term historical observation trajectories to extract spatial-temporal cues related to target positions and movement tendencies, providing complementary insights that boost the quality of otherwise ambiguous intermediate features. The module is capable of processing an increased number of historical trajectory frames while maintaining linear complexity.
To boost the quality of cross-agent global features, we devise a cross-agent collaborative prediction module that leverages historical global spatial-temporal trajectories to predict current complementary cues. This capability also enables the ego agent to seamlessly switch to "collaborative prediction mode" when it cannot promptly receive shared messages from neighboring agents. By independently predicting current global features, the model effectively mitigates the loss of complementary information and enhances feature quality in dynamic communication environments.

We conducted extensive experiments on simulated datasets and real-world datasets, including OPV2V and V2XSet, to validate the effectiveness and efficiency of our model. The contributions of this paper are summarized as follows:

\begin{itemize}
\item To the best of our knowledge, CollaMamba is the first unified framework in multi-agent collaborative perception which leverages Mamba and replaces traditional CNN and Transformer architectures. It produces compact, comprehensive sequence-form intermediate feature representations with linear complexity, significantly reducing computational and communication overhead.
\item We design history-aware feature boosting modules that can process and leverage long-term historical spatial-temporal sequences to extract complementary cues, refining and boosting vague target features at the current moment, thereby improving feature quality.
\item We devise a cross-agent global feature boosting method that improves the ego agent's global perception accuracy even in the absence of collaboration through collaborative prediction, allowing the model to adapt to dynamic real-world communication conditions.
\end{itemize}

\section{Related Work}
\textbf{Collaborative Perception} leverages the shared sensory data among multiple agents to enhance perception accuracy, robustness, and range. Current methodologies are primarily divided into early, intermediate, and late fusion techniques. Intermediate fusion is particularly preferred due to its effective compromise between performance and transmission bandwidth \cite{yazganSurveyIntermediateFusion2024,hanCollaborativePerceptionAutonomous2023}. 
In DiscoNet, \cite{liLearningDistilledCollaboration2021a} presented a distilled collaboration graph that models adaptive, pose-aware collaboration among agents using knowledge distillation.
In V2X-ViT, \cite{xuV2XViTVehicletoEverythingCooperative2022a} introduced a vision Transformer-based framework with multi-agent self-attention and multi-scale window self-attention. 
CoBEVT \cite{xuCoBEVTCooperativeBird2022} proposed a multi-agent, multi-camera perception framework for generating bird's-eye view (BEV) semantic segmentation maps, achieving state-of-the-art performance through efficient data fusion.
Where2comm\cite{huWhere2commCommunicationEfficientCollaborative2022} introduced a communication-efficient collaborative perception framework using spatial confidence maps, reducing communication overhead while maintaining high perception performance
In Select2Col, \cite{liuSelect2ColLeveragingSpatialTemporal2023} proposed a framework that leverages spatial-temporal importance of semantic information for efficient collaborative perception using a lightweight graph neural network and hybrid attention mechanism.
In HM-ViT, \cite{xiangHMViTHeteromodalVehicletoVehicle2023} introduced a hetero-modal cooperative perception framework using vision transformers, effectively fusing features from multi-view images and LiDAR point clouds. 
\cite{luExtensibleFrameworkOpen2024} proposed HEAL that accommodates heterogeneous agents, using pyramid structured network to fuse features and aligning new agents to a unified feature space.
Unlike previous methods that utilize either CNNs or Transformers to produce two-dimensional features laden with redundant information, our approach extracts more compact features in a sequential form. By integrating longer historical trajectory sequences for feature enhancement, our method demonstrates superior efficiency and effectiveness in collaborative feature fusion.

\noindent \textbf{State Space Model} (SSMs) have gained significant attention for their ability to capture long-range dependencies with linear complexity in various domains. The Mamba architecture, a recent advancement in SSMs, has demonstrated remarkable performance across multiple applications.  
\cite{guMambaLinearTimeSequence2023} first presented improvements to state space models for sequence modeling, achieving state-of-the-art performance in language, audio, and genomics with better efficiency than Transformers. 
\cite{ruanVMUNetVisionMamba2024} proposed a bidirectional state space model for visual representation learning, demonstrating higher performance and efficiency compared to vision transformers.  
\cite{liCFMWCrossmodalityFusion2024} proposed the Cross-modality Fusion Mamba with Weather-removal for multispectral object detection under adverse weather conditions. 
\cite{liVideoMambaStateSpace2024} adapted the Mamba architecture for video understanding, addressing local redundancy and global dependencies. 
\cite{tengDiMDiffusionMamba2024} proposed Diffusion Mamba for efficient high-resolution image synthesis, integrating Mamba with diffusion models to improve performance.  
Our method is the first to employ SSMs in the realm of multi-agent collaborative perception. It shares more compact and comprehensive features, thereby enhancing the capabilities of feature fusion and long-range sptial-temporal modeling.

\section{Method}

\begin{figure*}[]
    \centering
    \includegraphics[width=0.8\textwidth]{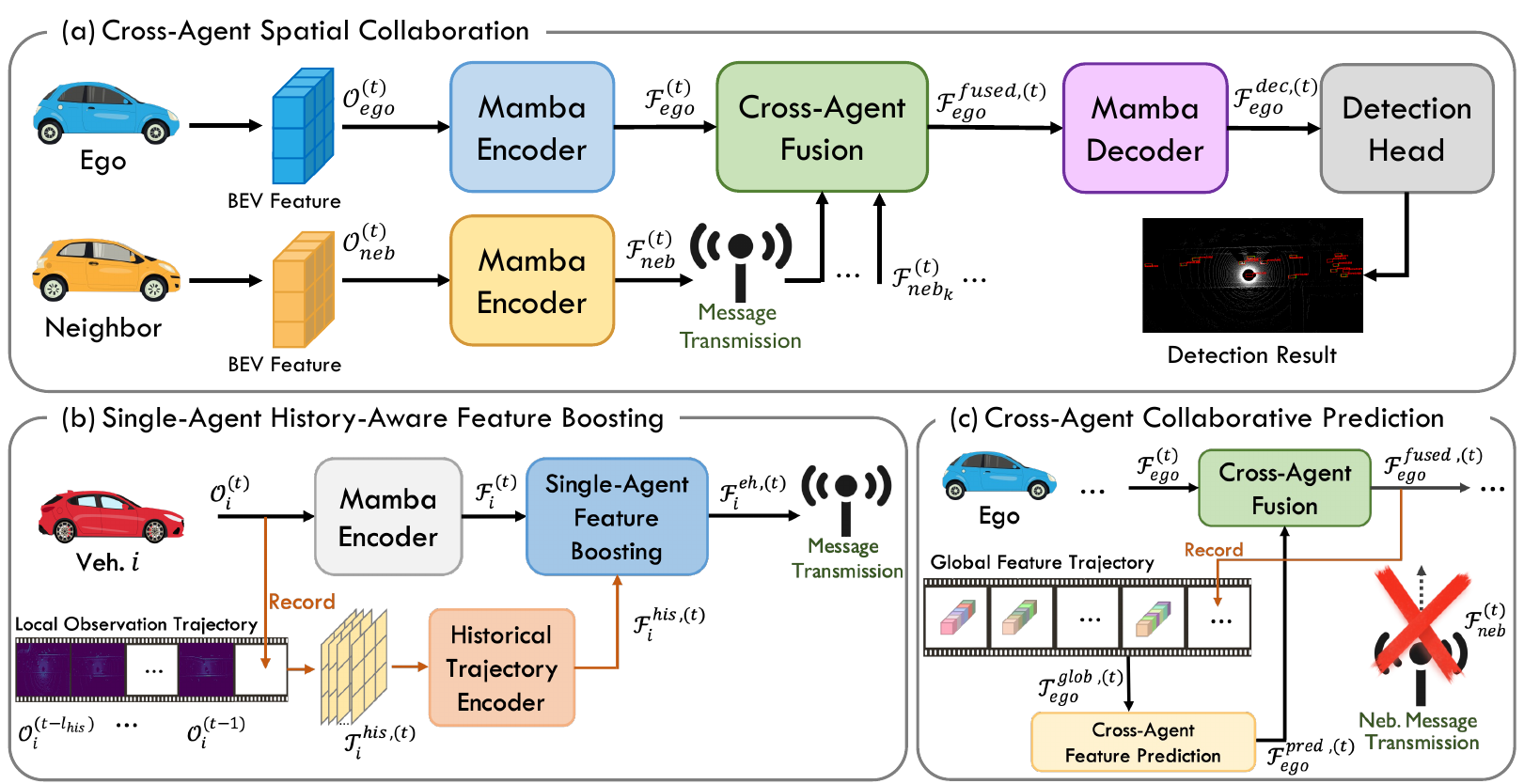}
    \caption{The framework of CollaMamba. (a) The structure of Cross-Agent Spatial Collaboration backbone network, yielding compact and comprehensive intermediate features; (b) Single-Agent History-Aware Feature Boosting module, extracting temporal contextual cues from local observation trajectory; (c) Cross-Agent Collaborative Prediction module, leveraging the broad spatial-temporal features extracted from the Global Feature Trajectory to predict missing information from neighbor agents.}
    \label{fig-overview}
\end{figure*}

\subsection{Cross-Agent Spatial Collaboration}

We first introduce the foundational backbone for cross-agent spatial collaboration framework based on SSMs, called CollaMamba-Simple, as shown in Figure \ref{fig-overview}(a). This framework primarily consists of three main modules: Mamba Encoder, Cross-Agent Fusion, and Mamba Decoder. These modules move away from traditional CNN and Transformer architectures, offering linear complexity. This allows for efficient modeling of long-range causal dependencies with a large receptive field while improving operational efficiency.

The main process of Cross-Agent Spatial Collaboration is described as Equation (\ref{eq_simple}). At each time \(t\), each agent \(i\) first utilizes the Mamba Encoder to extract local spatial features from the BEV observation \(\mathcal{O}^{(t)}_i\), resulting in a compact sequence-form intermediate feature \(\mathcal{F}^{(t)}_i\). Subsequently, agents share messages among themselves. Upon receiving messages from neighbor agents, the ego agent employs the Cross-Agent Fusion module to integrate its intermediate features \(\mathcal{F}^{(t)}_{ego}\) with the intermediate features \(\mathcal{F}^{(t)}_{neb}\) from the neighbor agents, yielding the fused cross-agent features \(\mathcal{F}^{fused,(t)}_{ego}\), which provides a comprehensive description of the global perspective in a collaboration. Following this, the features are decoded by the Mamba Decoder to reconstruct two-dimensional spatial features \(\mathcal{F}^{dec,(t)}_{ego}\), making them compatible with various object detection heads, and ultimately producing the final object detection results $\mathcal{\hat{Y}}_i^{(t)}$. 

\begin{subequations}
\begin{align}
	& \mathcal{F}_{i}^{(t)} = f_{encoder}\left( \mathcal{O}_{i}^{(t)} \right), \label{eq_simple_a} \\
	& \mathcal{F}_{ego}^{fused,(t)} = f_{global\_fusion}\left( \mathcal{F}_{ego}^{(t)}, \mathcal{F}_{neb_{j}}^{(t)}, \mathcal{F}_{neb_{k}}^{(t)}, ...  \right), \label{eq_simple_b} \\
	& \mathcal{F}_i^{dec, (t)} = f_{decoder}\left( {F}_{ego}^{fused,(t)}\right) ,  \label{eq_simple_c} \\
	& \mathcal{\hat{Y}}_i^{(t)} = f_{head}\left( \mathcal{F}_i^{dec, (t)}  \right) .  \label{eq_simple_d}
\end{align}
\label{eq_simple}
\end{subequations}

\begin{figure*}[!htb]
    \centering
    \includegraphics[width=0.8\textwidth]{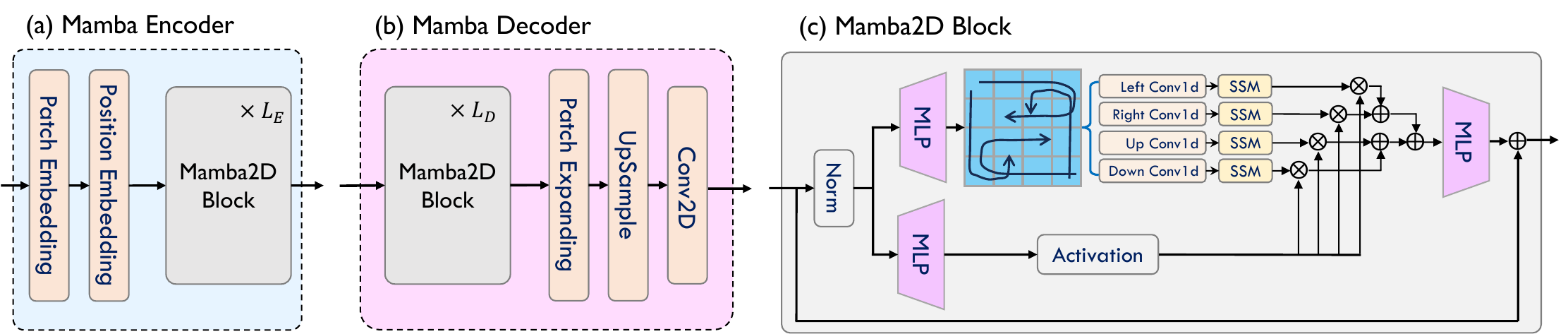}
    \caption{The structure of Mamba Encoder and Decoder module. (a) The structure of Mamba Encoder; (b) The structure of Mamba Decoder, for reconstructing spatial features, the scale and dimensions are adjusted after the Mamba2D blocks to accommodate the detection head; (c) Mamba2D Block, 4-direction SSM scan is employed to better extracting the compact sequence-form spatial casual dependencies.}
    \label{fig-enc-dec}
\end{figure*}

\noindent \textbf{Mamba Encoder and Decoder. \quad } Figure \ref{fig-enc-dec} illustrates the structure of the Mamba Encoder and Decoder modules. In the Mamba Encoder module (Figure \ref{fig-enc-dec}(a)), \(\mathcal{O}^{(t)}_i \in \textbf{R}^{b \times h_0 \times w_0 \times c_0 } \) is served as the input, where \(b\), \(h_0\), \(w_0\) and \(c_0\),  denote the batch-size, the height, width and embedding dimension of BEV feature map, respectively. This process extracts the single-agent local spatial features in sequential forms \(\mathcal{F}_i^{(t)} \in \textbf{R}^{b \times l \times c}\), where \(l\) and \(c\) denote the length and the embedding dimension of the feature sequence. Specifically, the two-dimensional BEV observation undergoes a Patch Embedding operation, followed by Position Embedding, and is finally fed into \(L_E\) serially connected Mamba2D modules. 


In every Mamba2D module, the spatial characteristics are captured by scanning the two-dimensional patches in four directions—left, right, up, and down—using SSMs to derive richer spatial relational features. In the context of the BEV view, each patch represents observational information within a certain spatial region. The relationships between patches indicate a connection among small observation areas, encompassing the relative positional relationship between the targets to be detected, the associations within cross-agent-regions of the same target, and the interactions between the target and its surrounding environment. The output of the Mamba Encoder is a compact and comprehensive sequential intermediate feature that describes both long-range spatial characteristics and deep semantic information with streamlined feature channels.

In the Mamba Decoder module (Figure \ref{fig-enc-dec}(b)), the input is intermediate features \(\mathcal{F}_i^{(t)}\), which are decoded and restored to two-dimensional spatial features \(\mathcal{F}_i^{detect, (t)}\) to meet the input requirements of the detection head. Initially, the intermediate features pass through \(L_D\) Mamba2D modules, followed by Patch Expanding, Upsampling, and Convolution2D operations. These operations adjust the height, width, and channel count of the feature maps to match the specifications of the object detection head.



\begin{figure}[!htb]
    \centering
    \includegraphics[width=0.4\textwidth]{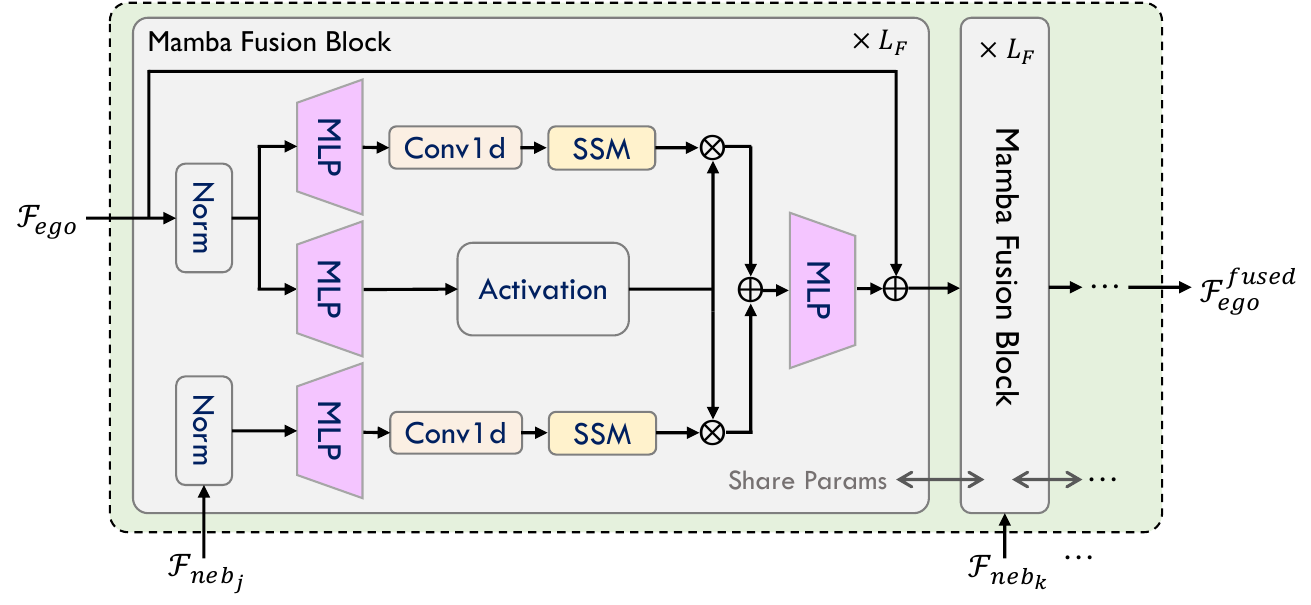}
    \caption{The structure of Cross-Agent Fusion module. }
    \label{fig-global-fusion}
\end{figure} 

\noindent \textbf{Cross-Agent Feature Fusion.\quad } Figure \ref{fig-global-fusion} illustrates the structure of the cross-agent feature fusion module, which effectively integrates the intermediate features of the ego agent with those of the neighbor agents. This module extracts cross-agent causal dependencies from the same spatial locations but different perspectives, or utilize complementary global spatial information to fill in the missing parts of the ego agent's field of view, thereby providing a clearer and more comprehensive description of the target objects within the global region.

For each neighbor agent around ego agent, the transmitted features are fused with the ego agent's features through a group of Mamba Fusion Block. Within the Mamba Fusion Block, the feature sequences of agents undergo SSM scanning to capture long-range mutual spatial dependencies. Multiple Mamba Fusion Blocks are employed, with shared parameters, of multiple neighbor agents. This approach enhances the model's scalability, allowing it to handle a dynamically changing number of neighbor agents effectively with linear complexity while efficiently integrating global features.

\noindent \textbf{Loss function.\quad } To train the proposed CollaMamba, we follow the previous works and employ the detection loss:

\begin{equation}
\mathcal{L} = \mathbf{E}_{i, t} \left[ L_{det}\left(  \mathcal{\hat{Y}}_i^{(t)}, \mathcal{Y}_i^{(t)} \right)  \right]
,
\label{eq_loss}
\end{equation}
where $ \mathcal{\hat{Y}}_i^{(t)}$ represents the output result, $\mathcal{Y}_i^{(t)}$ represenmts the corresponding ground-truth. The loss function $L_{det}(\cdot)$ of the model consists of a classification loss and a regression loss \cite{chenMilestonesAutonomousDriving2023}.  The training method and parameters are consistent with the SOTA works, and more details can be found in the appendix.


\subsection{Single-Agent History-Aware Feature Boosting}

%
Historical trajectory sequences in object tracking and detection offer vital information regarding object movement and motion tendency cues, thereby enhancing the accuracy of target localization at the current moment \cite{liSpatiotemporalFusionRemote2020, chenTrajectoryFormer3DObject2023}. Additionally, these sequences can provide complementary temporal information, as targets obscured in the current timestamp might have been visible in previous ones. Motivated by these insights, we have extended the CollaMamba-Simple framework by incorporating single-agent feature boosting module, resulting in a novel framework for cross-agent spatial-temporal collaboration, designated as CollaMamba-ST. This framework leverages the feature encoding of historical trajectory sequences to boost and refine the vague features in the present moment. In CollaMamba-ST, each single agent maintains a local observation trajectory to store historical observation data from the local perspective. Additionally, a Single-Agent Feature Boosting module is added after the Mamba Encoder of each agent, as illustrated in Figure \ref{fig-overview}(b). This module utilizes the long-term spatial-temporal features \(\mathcal{T}_i^{(t)}\) as hints and auxiliary information to refine and enhance the intermediate features \(\mathcal{F}_i^{(t)}\), resulting in the enhanced features \(\mathcal{F}_i^{eh, (t)}\). The spatial-temporal features of \(\mathcal{T}_i^{(t)}\) are extracted by an efficient historical trajectory encoder from locally cached historical trajectory sequences over an extended time period.

\begin{figure}[!htb]
    \centering
    \includegraphics[width=0.5\textwidth]{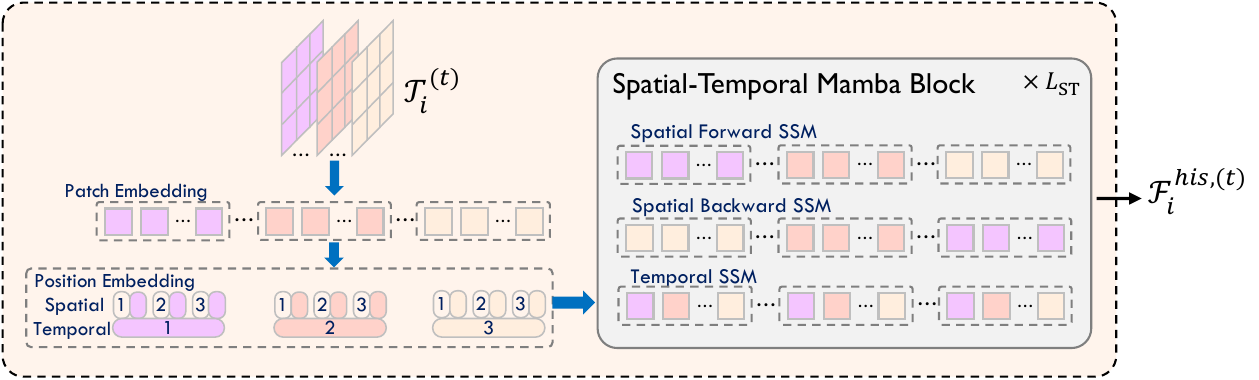}
    \caption{The structure of Historical Trajectory Encoder module. }
    \label{fig-ST-enc}
\end{figure} 

\noindent \textbf{Historical Trajectory Encoder.\quad } It is noteworthy that in encoder architectures utilizing the vanilla Transformer, multi-head attention exhibits quadratic complexity with respect to the number of tokens. This complexity is particularly relevant for long feature sequences, as the number of tokens increases with the number of input frames. Considering the efficiency and quality advantages of the Mamba in handling long sequences, we designed a Mamba-based historical trajectory encoder as shown in Figure \ref{fig-ST-enc}. 

During cross-agent spatial-temporal collaboration, each agent maintains a historical trajectory queue to cache historical observation feature maps with a length of \(l_{his}\), that is, the historical trajectory from timestep \(t - l_{his}\) to timestep \(t-1\). In historical trajectory encoder, longer historical observation feature  sequences \(\mathcal{T}_i^{(t)} \in \textbf{R}^{b, l_{his}, h_0, w_0, c_0}\) can be used as input to obtain the spatial-temporal feature encoding \(\mathcal{F}^{his}_i \in \textbf{R}^{b, l_{his}, l, c}\) of the historical trajectory sequence, as described in Equation (\ref{eq_his_enc}):
\begin{equation}
\begin{aligned}
& \mathcal{F}^{his, (t)}_i = f_{his\_encoder}\left(  \mathcal{T}_i^{(t)}  \right), \\
& \mathcal{T}_i^{(t)} = stack\left(\left[\mathcal{O}_{i}^{bev, (t - l_{his})}, ..., \mathcal{O}_{i}^{bev, (t - 1)}\right]\right).
\end{aligned}
\label{eq_his_enc}
\end{equation}

Specifically, spatial position embedding and temporal position embedding are applied to the historical sequence \(\mathcal{T}\). Then, it is fed into \(L_{ST}\) consecutive  Spatial-Temporal Mamba Blocks for spatial-temporal feature encoding. In each Spatial-Temporal Mamba Block, considering that the primitive historical spatiotemporal features differ from texts in that they amass non-causal 2D spatial information in addition to temporal redundant information, we further designed a parallel three-directional SSM scanning method to address this issue of adapting to non-causal input. Specifically, each frame's 2D feature map is unfolded into a 1D sequence along rows and columns, and then the frame sequences are concatenated front-to-back, yielding the sequence \(H^s_i \in \textbf{R}^{b, l_{his}(h_0w_0), c}\), which undergoes forward SSM scanning. Additionally, backward SSM scanning is performed to capture more detailed spatial positional relationships and causal dependencies without significantly increasing computational complexity. Meanwhile, patches are stacked along the temporal dimension to construct the sequence \(H^t_i \in \textbf{R}^{b, (h_0w_0)l_{his}, c}\), which undergoes temporal SSM scanning to capture the temporal variations and causal relationships at the same spatial position. Consequently, this yields a spatial-temporal sequence feature rich in intrinsic temporal and spatial dependencies.

\begin{figure}[!htb]
    \centering
    \includegraphics[width=0.35\textwidth]{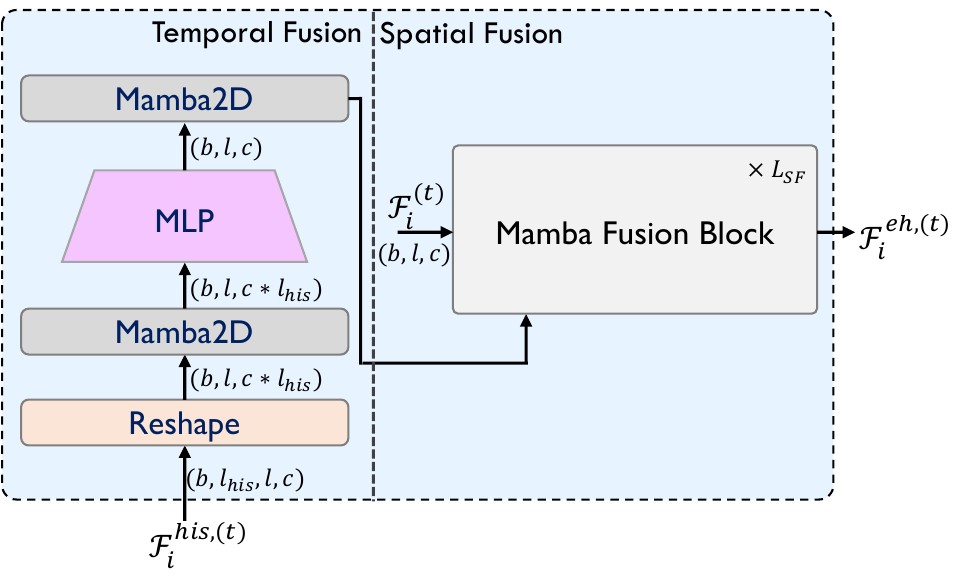}
    \caption{The structure of Single-Agent Feature Boosting module.}
    \label{fig-local-enhancer}
\end{figure} 
\noindent \textbf{Single-Agent Feature Boosting. \quad }  As shown in Equation (\ref{eq_eh}), The spatial-temporal sequence features \(\mathcal{F}^{his,(t)}_i\), which encompass contextual information about the position and motion state of the target to be detected, are employed for feature information complementation across temporal scales. These features are used to correct and refine the local features \(\mathcal{F}_i^{(t)}\), resulting in the enhanced intermediate features \(\mathcal{F}_i^{eh,(t)}\).
\begin{equation}
\mathcal{F}_i^{eh, (t)} = f_{boost}\left(  \mathcal{F}_i^{(t)}, \mathcal{F}^{his, (t)}_i \right).
\label{eq_eh}
\end{equation}

The design details of the single-agent feature boosting module are illustrated in Figure \ref{fig-local-enhancer}. Initially, \(\mathcal{F}^{his,(t)}_i\) undergoes temporal fusion and dimensionality reduction (Figure \ref{fig-local-enhancer} left). This process concatenates the features along the temporal dimension, and employs a Mamba2D Block to fuse the temporal features at each spatial position, followed by dimensionality reduction through MLP, yielding the auxiliary information at each spatial location provided by the historical trajectory. Subsequently, the local intermediate features are spatially fused with this auxiliary information sequence (Figure \ref{fig-local-enhancer} right). This fusion process continues to employ the network of Mamba Fusion Block described in the previous section.

\subsection{Cross-Agent Collaborative Prediction}
In practical vehicular network communication scenarios, the increase in the number of collaborative agents and the dynamic changes in channel conditions make communication resources extremely limited. At certain times, when receiving intermediate features from neighboring agents, high delays, untimely reception, or even complete failure to receive are inevitable. If the ego agent fails to receive messages from neighboring agents in a timely manner, it cannot afford to wait indefinitely. Instead, it must promptly process its local intermediate features through the subsequent decoder and detection head to ensure timely and effective target detection results. This lack of global collaborative information may lead to reduced accuracy. 

To address this issue of poor neighbor communication, we aim to compensate to some extent by using historical global features to predict the current collaborative features. Consequently, we have developed a history-aware cross-agent collaborative prediction module that can be effortlessly integrated into the CollaMamba backbone as a plug-and-play component. This enhancement empowers the model to adaptively predict missing neighbor information and global features, giving rise to the CollaMamba-Miss, as illustrated in Figure \ref{fig-overview}(c). The cross-agent collaborative prediction module includes a global feature trajectory sequence buffer to retain historical global features and a global feature prediction module to anticipate missing collaborative information, thereby enhancing the global intermediate features.
This design enables model to seamlessly switch between the ``Neighbor Feature Fusion mode" and the ``Collaborative Prediction mode". For ego agent, if the features from neighboring agents are received within the specified time, cross-agent feature fusion is performed as described in CollaMamba-Simple or CollaMamba-ST. If the specified time is exceeded and the collaborative information from neighboring agents is still not will-received, the system switches to the Collaborative Prediction mode, using historical global feature trajectory to predict the current global features, as illustrated in Equation (\ref{eq_global_pred}):
\begin{equation}
\begin{aligned}
& \mathcal{F}_{ego}^{fused, (t)} = f_{global\_fusion}\left( \mathcal{F}_{ego}^{(t)}, \mathcal{F}_{neb_{j}}^{(t)}, \mathcal{F}_{neb_{k}}^{(t)}, ...  \right),  \\
& if \quad \Delta \tau \leq \tau_0 \quad and \quad recv\_flag=true; \\
& \mathcal{F}_{ego}^{fused, (t)} = f_{global\_fusion}\left( \mathcal{F}_{ego}^{(t)}, f_{glob\_pred} \left( \mathcal{T}_{ego}^{glob,(t)} \right)\right),  \\
& \quad \mathcal{T}_{ego}^{glob,(t)} = stack\left(\left[\mathcal{F}_{ego}^{fused,(t-l_{his})}, ..., \mathcal{F}_{ego}^{fused,(t-1)}\right]\right), \\
& elif \quad \Delta \tau \textgreater \tau_0 \quad and \quad recv\_flag=false,
\end{aligned}
\label{eq_global_pred}
\end{equation}
where $\Delta \tau$ represents the waiting time of the ego agent, \(\tau_0\) represents the maximum allowed waiting time for receiving messages, $\mathcal{T}_{ego}^{glob,(t)}$ represents the spatial-temporal sequence composed of global feature trajectory. The global feature prediction module here, $f_{glob\_pred}\left( \cdot \right)$, is similar to the historical trajectory encoder module described in Figure \ref{fig-ST-enc}, with the only difference being that the input is now a historical trajectory sequence composed of one-dimensional global features.

\section{Experiments}


\begin{table*}[]
\centering
\caption{Comparison of performance and inference overhead on OPV2V and V2XSet datasets.}
\label{table-comparison}

\begin{tabular}{@{}c|c|ccc|ccc@{}}
\toprule
\multirow{2}{*}{Model} & \multirow{2}{*}{\begin{tabular}[c]{@{}c@{}}\#Params\\ (M)\end{tabular}} & \multicolumn{3}{c|}{OPV2V}       & \multicolumn{3}{c}{V2XSet}      \\ \cmidrule(l){3-8} 
                       &                                                                        & AP@0.5/0.7    & \#CV  & FLOPs(G) & AP@0.5/0.7    & \#CV & FLOPs(G) \\ \midrule
AttFuse                & 8.06                                                                   & 0.923 / 0.796 & 25.7  & 100.01   & 0.848 / 0.669 & 25.7 & 104.33   \\
V2VNet                 & 14.61                                                                  & 0.945 / 0.811 & 25.7  & 496.65   & 0.887 / 0.658 & 25.7 & 531.55   \\
V2X-ViT                & 13.45                                                                  & 0.963 / 0.872 & 25.6  & 280.96   & 0.904 / 0.751 & 25.6 & 285.09   \\
CoBEVT                 & 10.51                                                                  & 0.931 / 0.719 & 25.6  & 207.26   & 0.894 / 0.727 & 25.6 & 211.40   \\
How2comm                 & 35.79                                                                  & 0.854 / 0.722 & 25.6  & 949.82   & 0.841 / 0.670 & 25.6 & 741.34   \\
Where2comm-single      & 11.43                                                                  & 0.894 / 0.801 & 20.96 & 111.92   & 0.806 / 0.582 & 22.2 & 116.77   \\
Where2comm-multi       & 11.43                                                                  & 0.946 / 0.807 & 22.94 & 218.90   & 0.894 / 0.736 & 20.2 & 228.49   \\ \midrule
CollaMamba-Simple      & \textbf{3.92}                                                                   & 0.947 / 0.892 & \textbf{19.7}  & \textbf{79.06}    & 0.918 / 0.765 & \textbf{19.7} & \textbf{75.88}    \\
	CollaMamba-ST          & 5.35                                                                   & \textbf{0.975 / 0.913 }& \textbf{19.7}  & 107.87   &  \textbf{0.932 / 0.798}             & \textbf{19.7} & 104.67         \\
CollaMamba-Miss        & 6.25                                                                   & 0.972 / 0.902 & \textbf{19.7}  & 111.87   &  0.931 / 0.778             & \textbf{19.7} & 108.68         \\ \bottomrule
\end{tabular}
\end{table*}

\subsection{Datasets and Experiment Settings}
We perform comprehensive evaluations on serveral benchmark datasets, including OPV2V \cite{xuOPV2VOpenBenchmark2022a}, V2XSet \cite{xuV2XViTVehicletoEverythingCooperative2022a}. The experiments are conducted on RTX 4090 GPU, and the models are implemented using PyTorch 2.1. Following \cite{huWhere2commCommunicationEfficientCollaborative2022}, we evaluate 3D object detection performance using Average Precision (AP) at IoU thresholds of 0.5 and 0.7. The number of model parameters, denoted as ``\#params", is measured in millions (M). Communication volume, referred to as Communication-Volume (\#CV), represents the size of messages transmitted by neighbor agents, measured in bytes and expressed as a base-2 logarithm. ``FLOPs" is used to measure the computational complexity of the model. Detailed descriptions of the dataset, evaluation metrics, experimental environment, and parameter settings are provided in the appendix.

\subsection{Comparison Analysis}

To validate the effectiveness of our model, we selected the following baseline models for comparison: AttFuse \cite{xuOPV2VOpenBenchmark2022b}, V2VNet \cite{wangV2VNetVehicletoVehicleCommunication2020}, V2X-ViT \cite{huWhere2commCommunicationEfficientCollaborative2022}, CoBEVT \cite{xuCoBEVTCooperativeBird2022}, Where2comm \cite{huWhere2commCommunicationEfficientCollaborative2022}, How2comm \cite{yangHow2commCommunicationEfficientCollaborationPragmatic2023}. ``Where2Comm-single" refers to the use of a single-scale fusion method, while ``Where2Comm-multi" refers to the use of a multi-scale fusion method. 
``CollaMamba-Simple" refers to the lightweight basic backbone framework proposed in this paper; "CollaMamba-ST" denotes the spatio-temporal CollaMamba model with the single-agent history-aware feature boosting module; and ``CollaMamba-Miss" represents the version with the cross-agent collaborative prediction module, designed to handle scenarios with poor neighbor communication. The length of the historical trajectory sequence buffer \(l_{his}\) is set to 10 for CollaMamba-ST and 20 for CollaMamba-Miss.
 The comparison of prediction accuracy for each model is shown in Table \ref{table-comparison}.
 
 From the results presented in Table \ref{table-comparison}, it is evident that on the OPV2V dataset, the proposed CollaMamba models significantly reduce communication overhead compared to the representative Transformer-based method V2X-ViT. Specifically, communication volume is reduced to approximately $1/64$ across the three CollaMamba models. The lightweight CollaMamba-Simple framework achieves a 70.9$\%$  reduction in the number of parameters and a 71.9$\%$ reduction in FLOPs. CollaMamba-ST reduces parameters by 60.2$\%$  and FLOPs by 61.6$\%$ , while CollaMamba-Miss shows a 53.5$\%$ reduction in parameters and a 60.2$\%$  reduction in FLOPs. 
In terms of accuracy, the three CollaMamba models maintain, or even improve, performance, with CollaMamba-ST achieving a 4.1$\%$  increase in AP@0.7, attributed to the contextual reference provided by the historical sequence buffer.

 When compared to the communication-efficient Where2comm-multi method, CollaMamba models reduce communication volume to $1/9.45$, while significantly enhancing accuracy. CollaMamba-ST, for example, improves AP@0.7 by 10.6$\%$ , reduces parameters by 53.2$\%$, and decreases FLOPs by 50$\%$ . CollaMamba-Simple sees an 8.5$\%$  increase in AP@0.7, with a 65.7$\%$  reduction in parameters and a 63.9$\%$  decrease in FLOPs. CollaMamba-Miss also shows a 9.5$\%$  improvement in AP@0.7, with a 45.3$\%$  reduction in parameters and a 48.9$\%$  decrease in FLOPs. It is worth noting that the How2comm method, although capable of enhancing model performance by incorporating one or two frames of historical information as auxiliary cues through its complex temporal attention mechanism, results in significantly high FLOPs. Additional experimental results can be found in the appendix.

Overall, compared to other methods, the CollaMamba models significantly reduce both communication and computational overhead while maintaining, or even improving, accuracy. The superior performance of the CollaMamba models is largely due to effective modeling of long-range spatial dependencies and the utilization of long-term historical spatio-temporal sequences to boost intermediate features.

For the component ablation study, the results in Table \ref{table-comparison} also show that CollaMamba-ST and CollaMamba-Miss surpass CollaMamba-Simple in terms of model accuracy. This indicates that the history-aware feature boosting modules we designed effectively enhances model performance, demonstrating that both single-agent and cross-agent historical trajectory spatial-temporal modeling and feature enhancement strategies can improve the quality of intermediate features.

\subsection{Analysis of Poor Neighbor Communication}

\begin{figure}[!htb]
    \centering
    \includegraphics[width=0.5\textwidth]{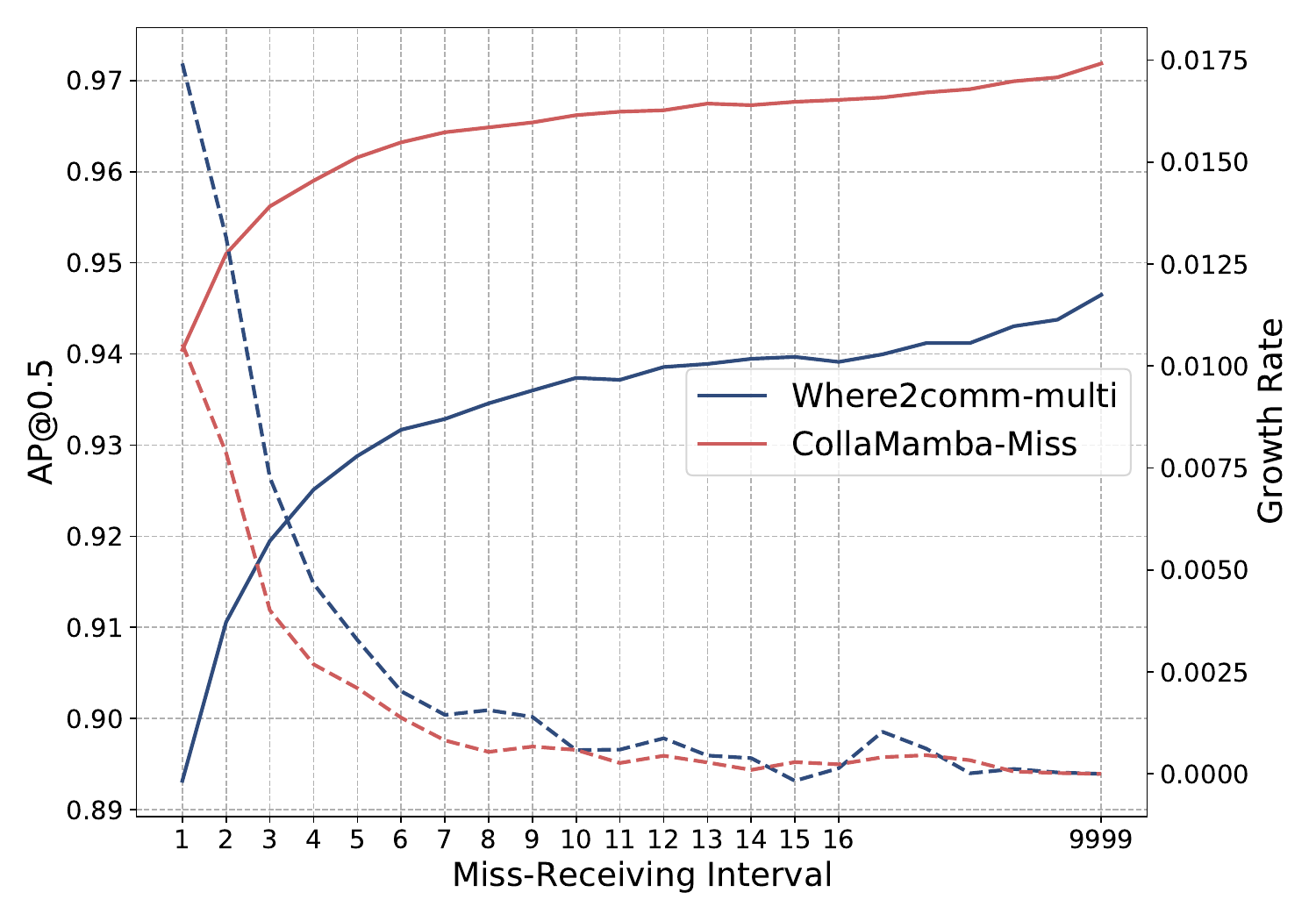}
    \caption{Robustness to poor neighbor communication, the solid line depicts the change in AP@0.5, while the dashed line illustrates the accuracy growth rate.}
    \label{fig-poor}
\end{figure}

To simulate real-world scenarios where the ego agent may occasionally fail to receive auxiliary information from neighbor agents due to communication quality issues, we designed the Miss-Receiving experiment, as illustrated in Figure \ref{fig-poor}. The x-axis represents the interval at which miss-receiving occurs. A larger miss-receiving interval indicates a lower probability of poor neighbor communication and miss-receiving events, meaning the ego agent can receive auxiliary information from Neighbor agents more frequently.

For the Where2comm method, accuracy improves significantly as the miss-receiving interval increases, particularly when the interval is between 1 and 5. This suggests that the model's performance is highly dependent on more frequent transmissions and assistance from neighbor agents. In contrast, CollaMamba-Miss achieves relatively high accuracy even when miss-receiving occurs every other frame. As the miss-receiving interval increases, the accuracy improvement slows, suggesting that during miss-receiving events, CollaMamba-Miss can rely on its cross-agent collaborative perception module to partially compensate for the missing global features using the spatial-temporal characteristics of historical trajectories.
This experiment highlights the robustness of CollaMamba-Miss, which, owing to its long-term spatiotemporal trajectory modeling capabilities, effectively compensates for missing features during miss-receiving events by leveraging motion cues from historical moments and adapts to varying miss-receiving intervals in challenging neighbor communication scenarios.

\section{Conclusion}
In this paper, we introduce CollaMamba, a cross-agent collaborative perception architecture that efficiently models long-range spatial-temporal dependencies while maintaining linear complexity and reducing computational and communication overhead. The history-aware feature boosting module enhances ambiguous features by processing long-term historical trajectory sequences, while the cross-agent collaborative prediction module allows the model to adapt to dynamically changing communication scenarios. Extensive experiments on various datasets demonstrate the effectiveness of our model and the necessity of its components. Future research will focus on enhancing feature processing for long-term point cloud sequences and improving multimodal fusion in complex scenarios.



\bibliography{aaai25}

\begin{thebibliography}{26}
\providecommand{\natexlab}[1]{#1}

\bibitem[{Chen et~al.(2023{\natexlab{a}})Chen, Li, Huang, Li, Xing, Tian, Li, Hu, Na, Li, Teng, Lv, Wang, Cao, Zheng, and Wang}]{chenMilestonesAutonomousDriving2023}
Chen, L.; Li, Y.; Huang, C.; Li, B.; Xing, Y.; Tian, D.; Li, L.; Hu, Z.; Na, X.; Li, Z.; Teng, S.; Lv, C.; Wang, J.; Cao, D.; Zheng, N.; and Wang, F.-Y. 2023{\natexlab{a}}.
\newblock Milestones in {{Autonomous Driving}} and {{Intelligent Vehicles}}: {{Survey}} of {{Surveys}}.
\newblock \emph{IEEE Transactions on Intelligent Vehicles}, 8(2): 1046--1056.

\bibitem[{Chen et~al.(2023{\natexlab{b}})Chen, Wu, Chitta, Jaeger, Geiger, and Li}]{chenEndtoendAutonomousDriving2023}
Chen, L.; Wu, P.; Chitta, K.; Jaeger, B.; Geiger, A.; and Li, H. 2023{\natexlab{b}}.
\newblock End-to-End {{Autonomous Driving}}: {{Challenges}} and {{Frontiers}}.

\bibitem[{Chen et~al.(2023{\natexlab{c}})Chen, Shi, Zhang, Zhu, Wang, Cheung, See, and Li}]{chenTrajectoryFormer3DObject2023}
Chen, X.; Shi, S.; Zhang, C.; Zhu, B.; Wang, Q.; Cheung, K.~C.; See, S.; and Li, H. 2023{\natexlab{c}}.
\newblock {{TrajectoryFormer}}: {{3D Object Tracking Transformer}} with {{Predictive Trajectory Hypotheses}}.
\newblock arXiv:2306.05888.

\bibitem[{Dosovitskiy et~al.(2017)Dosovitskiy, Ros, Codevilla, Lopez, and Koltun}]{dosovitskiyCARLAOpenUrban2017}
Dosovitskiy, A.; Ros, G.; Codevilla, F.; Lopez, A.; and Koltun, V. 2017.
\newblock {{CARLA}}: {{An Open Urban Driving Simulator}}.
\newblock arXiv:1711.03938.

\bibitem[{Gao et~al.(2024)Gao, Zhang, Lu, Huang, Yang, Xiong, and Liu}]{gaoSurveyCollaborativePerception2024}
Gao, X.; Zhang, X.; Lu, Y.; Huang, Y.; Yang, L.; Xiong, Y.; and Liu, P. 2024.
\newblock A {{Survey}} of {{Collaborative Perception}} in {{Intelligent Vehicles}} at {{Intersections}}.
\newblock \emph{IEEE Transactions on Intelligent Vehicles}, 1--20.

\bibitem[{Gu and Dao(2023)}]{guMambaLinearTimeSequence2023}
Gu, A.; and Dao, T. 2023.
\newblock Mamba: {{Linear-Time Sequence Modeling}} with {{Selective State Spaces}}.
\newblock arXiv:2312.00752.

\bibitem[{Han et~al.(2023)Han, Zhang, Li, Jin, Lang, and Li}]{hanCollaborativePerceptionAutonomous2023}
Han, Y.; Zhang, H.; Li, H.; Jin, Y.; Lang, C.; and Li, Y. 2023.
\newblock Collaborative {{Perception}} in {{Autonomous Driving}}: {{Methods}}, {{Datasets}}, and {{Challenges}}.
\newblock \emph{IEEE Intelligent Transportation Systems Magazine}, 15(6): 131--151.

\bibitem[{Hu et~al.(2022)Hu, Fang, Lei, Zhong, and Chen}]{huWhere2commCommunicationEfficientCollaborative2022}
Hu, Y.; Fang, S.; Lei, Z.; Zhong, Y.; and Chen, S. 2022.
\newblock Where2comm: {{Communication-Efficient Collaborative Perception}} via {{Spatial Confidence Maps}}.
\newblock arXiv:2209.12836.

\bibitem[{Li et~al.(2024{\natexlab{a}})Li, Hu, Yao, Yang, and Chen}]{liCFMWCrossmodalityFusion2024}
Li, H.; Hu, Q.; Yao, Y.; Yang, K.; and Chen, P. 2024{\natexlab{a}}.
\newblock {{CFMW}}: {{Cross-modality Fusion Mamba}} for {{Multispectral Object Detection}} under {{Adverse Weather Conditions}}.
\newblock arXiv:2404.16302.

\bibitem[{Li et~al.(2020)Li, Li, He, Chen, and Plaza}]{liSpatiotemporalFusionRemote2020}
Li, J.; Li, Y.; He, L.; Chen, J.; and Plaza, A. 2020.
\newblock Spatio-Temporal Fusion for Remote Sensing Data: An Overview and New Benchmark.
\newblock \emph{Science China Information Sciences}, 63(4): 140301.

\bibitem[{Li et~al.(2024{\natexlab{b}})Li, Li, Wang, He, Wang, Wang, and Qiao}]{liVideoMambaStateSpace2024}
Li, K.; Li, X.; Wang, Y.; He, Y.; Wang, Y.; Wang, L.; and Qiao, Y. 2024{\natexlab{b}}.
\newblock {{VideoMamba}}: {{State Space Model}} for {{Efficient Video Understanding}}.

\bibitem[{Li et~al.(2021)Li, Ren, Wu, Chen, Feng, and Zhang}]{liLearningDistilledCollaboration2021a}
Li, Y.; Ren, S.; Wu, P.; Chen, S.; Feng, C.; and Zhang, W. 2021.
\newblock Learning {{Distilled Collaboration Graph}} for {{Multi-Agent Perception}}.
\newblock In \emph{Advances in {{Neural Information Processing Systems}}}, volume~34, 29541--29552. Curran Associates, Inc.

\bibitem[{Liu et~al.(2023)Liu, Huang, Li, Chen, Zhao, Zhao, Zhu, and Zhang}]{liuSelect2ColLeveragingSpatialTemporal2023}
Liu, Y.; Huang, Q.; Li, R.; Chen, X.; Zhao, Z.; Zhao, S.; Zhu, Y.; and Zhang, H. 2023.
\newblock {{Select2Col}}: {{Leveraging Spatial-Temporal Importance}} of {{Semantic Information}} for {{Efficient Collaborative Perception}}.
\newblock arXiv:2307.16517.

\bibitem[{Lu et~al.(2024)Lu, Hu, Zhong, Wang, Chen, and Wang}]{luExtensibleFrameworkOpen2024}
Lu, Y.; Hu, Y.; Zhong, Y.; Wang, D.; Chen, S.; and Wang, Y. 2024.
\newblock An {{Extensible Framework}} for {{Open Heterogeneous Collaborative Perception}}.
\newblock arXiv:2401.13964.

\bibitem[{Ruan and Xiang(2024)}]{ruanVMUNetVisionMamba2024}
Ruan, J.; and Xiang, S. 2024.
\newblock {{VM-UNet}}: {{Vision Mamba UNet}} for {{Medical Image Segmentation}}.
\newblock arXiv:2402.02491.

\bibitem[{Teng et~al.(2024)Teng, Wu, Shi, Ning, Dai, Wang, Li, and Liu}]{tengDiMDiffusionMamba2024}
Teng, Y.; Wu, Y.; Shi, H.; Ning, X.; Dai, G.; Wang, Y.; Li, Z.; and Liu, X. 2024.
\newblock {{DiM}}: {{Diffusion Mamba}} for {{Efficient High-Resolution Image Synthesis}}.
\newblock arXiv:2405.14224.

\bibitem[{Wang et~al.(2020)Wang, Manivasagam, Liang, Yang, Zeng, and Urtasun}]{wangV2VNetVehicletoVehicleCommunication2020}
Wang, T.-H.; Manivasagam, S.; Liang, M.; Yang, B.; Zeng, W.; and Urtasun, R. 2020.
\newblock {{V2VNet}}: {{Vehicle-to-Vehicle Communication}} for {{Joint Perception}} and {{Prediction}}.
\newblock In Vedaldi, A.; Bischof, H.; Brox, T.; and Frahm, J.-M., eds., \emph{Computer {{Vision}} -- {{ECCV}} 2020}, volume 12347, 605--621. Cham: Springer International Publishing.
\newblock ISBN 978-3-030-58535-8 978-3-030-58536-5.

\bibitem[{Xiang, Xu, and Ma(2023)}]{xiangHMViTHeteromodalVehicletoVehicle2023}
Xiang, H.; Xu, R.; and Ma, J. 2023.
\newblock {{HM-ViT}}: {{Hetero-modal Vehicle-to-Vehicle Cooperative Perception}} with {{Vision Transformer}}.
\newblock In \emph{2023 {{IEEE}}/{{CVF International Conference}} on {{Computer Vision}} ({{ICCV}})}, 284--295. Paris, France: IEEE.
\newblock ISBN 9798350307184.

\bibitem[{Xu et~al.(2021)Xu, Guo, Han, Xia, Xiang, and Ma}]{xuOpenCDAOpenCooperative2021}
Xu, R.; Guo, Y.; Han, X.; Xia, X.; Xiang, H.; and Ma, J. 2021.
\newblock {{OpenCDA}}: {{An Open Cooperative Driving Automation Framework Integrated}} with {{Co-Simulation}}.
\newblock In \emph{2021 {{IEEE International Intelligent Transportation Systems Conference}} ({{ITSC}})}, 1155--1162.

\bibitem[{Xu et~al.(2022{\natexlab{a}})Xu, Tu, Xiang, Shao, Zhou, and Ma}]{xuCoBEVTCooperativeBird2022}
Xu, R.; Tu, Z.; Xiang, H.; Shao, W.; Zhou, B.; and Ma, J. 2022{\natexlab{a}}.
\newblock {{CoBEVT}}: {{Cooperative Bird}}'s {{Eye View Semantic Segmentation}} with {{Sparse Transformers}}.
\newblock arXiv:2207.02202.

\bibitem[{Xu et~al.(2022{\natexlab{b}})Xu, Xiang, Tu, Xia, Yang, and Ma}]{xuV2XViTVehicletoEverythingCooperative2022a}
Xu, R.; Xiang, H.; Tu, Z.; Xia, X.; Yang, M.-H.; and Ma, J. 2022{\natexlab{b}}.
\newblock {{V2X-ViT}}: {{Vehicle-to-Everything Cooperative Perception}} with {{Vision Transformer}}.

\bibitem[{Xu et~al.(2022{\natexlab{c}})Xu, Xiang, Xia, Han, Li, and Ma}]{xuOPV2VOpenBenchmark2022a}
Xu, R.; Xiang, H.; Xia, X.; Han, X.; Li, J.; and Ma, J. 2022{\natexlab{c}}.
\newblock {{OPV2V}}: {{An Open Benchmark Dataset}} and {{Fusion Pipeline}} for {{Perception}} with {{Vehicle-to-Vehicle Communication}}.
\newblock In \emph{2022 {{International Conference}} on {{Robotics}} and {{Automation}} ({{ICRA}})}, 2583--2589. Philadelphia, PA, USA: IEEE.
\newblock ISBN 978-1-72819-681-7.

\bibitem[{Xu et~al.(2022{\natexlab{d}})Xu, Xiang, Xia, Han, Li, and Ma}]{xuOPV2VOpenBenchmark2022b}
Xu, R.; Xiang, H.; Xia, X.; Han, X.; Li, J.; and Ma, J. 2022{\natexlab{d}}.
\newblock {{OPV2V}}: {{An Open Benchmark Dataset}} and {{Fusion Pipeline}} for {{Perception}} with {{Vehicle-to-Vehicle Communication}}.
\newblock arXiv:2109.07644.

\bibitem[{Yang et~al.(2023)Yang, Yang, Wang, Liu, Xu, Yin, Zhai, and Zhang}]{yangHow2commCommunicationEfficientCollaborationPragmatic2023}
Yang, D.; Yang, K.; Wang, Y.; Liu, J.; Xu, Z.; Yin, R.; Zhai, P.; and Zhang, L. 2023.
\newblock How2comm: {{Communication-Efficient}} and {{Collaboration-Pragmatic Multi-Agent Perception}}.
\newblock \emph{Advances in Neural Information Processing Systems}, 36: 25151--25164.

\bibitem[{Yazgan et~al.(2024)Yazgan, Graf, Liu, Fleck, and Zoellner}]{yazganSurveyIntermediateFusion2024}
Yazgan, M.; Graf, T.; Liu, M.; Fleck, T.; and Zoellner, J.~M. 2024.
\newblock A {{Survey}} on {{Intermediate Fusion Methods}} for {{Collaborative Perception Categorized}} by {{Real World Challenges}}.
\newblock arXiv:2404.16139.

\bibitem[{Zhu et~al.(2023)Zhu, Peng, Li, Shen, Huang, and Lei}]{zhuModelingLongrangeDependencies2023}
Zhu, J.; Peng, B.; Li, W.; Shen, H.; Huang, Q.; and Lei, J. 2023.
\newblock Modeling {{Long-range Dependencies}} and {{Epipolar Geometry}} for {{Multi-view Stereo}}.
\newblock \emph{ACM Transactions on Multimedia Computing, Communications, and Applications}, 19(6): 1--17.

\end{thebibliography}

\clearpage
\appendix
\begin{center}
    \textbf{\LARGE Appendix}
\end{center}

\section{A.1 Preliminaries}

The state space model (SSM) and Mamba derive inspiration from linear time-invariant systems. The primary objective is to map a one-dimensional function or sequence, denoted as $ x(t) \in \textbf{R} $, to $ y(t) $ via a hidden state $ h(t) \in \textbf{R}^N $. Within this framework, the matrix $ \textbf{A} \in \textbf{R}^{N \times N} $ functions as the system’s evolution parameter, while $ \textbf{B} \in \textbf{R}^{N \times 1} $,  $ \textbf{C} \in \textbf{R}^{1 \times N} $ and $ \textbf{D} \in \textbf{R}^{1} $ serve as the projection parameters. The system dynamics can be mathematically encapsulated by the following equation:		
\begin{equation}
\begin{aligned}
	& h'(t)=\textbf{A}h(t)+\textbf{B}x(t), \\
	& y(t)=\textbf{C}h(t)+\textbf{D}x(t),
\end{aligned}
\end{equation}
where $ \textbf{D} $ also represents the residual connection. The continuous ordinary differential equation (ODE) inherent in this model is approximated through discretization in contemporary SSMs. Mamba \cite{guMambaLinearTimeSequence2023} exemplifies a discrete version of the continuous system, incorporating a timescale parameter $ \Delta $ to convert the continuous parameters $ A $ and $ B $ to their discrete equivalents $\bar{\textbf{A}}$ and $\bar{\textbf{B}}$. This transformation typically employs the zero-order hold (ZOH) method, which is defined by the following equation:
\begin{equation}
\begin{aligned}
	& \bar{\textbf{A}} = \exp(\Delta \textbf{A}), \\
	& \bar{\textbf{B}} = (\Delta \textbf{A})^{-1} (\exp(\Delta \textbf{A}) - \textbf{I}) \cdot \Delta \textbf{B} ,
\end{aligned}
\end{equation}

Consequently, the discrete representation of this linear system can be formulated in the following recurrent form:
\begin{equation}
\begin{aligned}
	& h_t=\bar{\textbf{A}}h_{t-1}+\bar{\textbf{B}}x_t, \\
	& y_t=\textbf{C}h_t+\textbf{D}x_t,
\end{aligned}
\end{equation}

Finally, the output is derived through global convolution, which be used for efficient parallel training:
\begin{equation}
\begin{aligned}
&\bar{\mathbf{K}}=(\mathbf{C}\bar{\mathbf{B}},\mathbf{C}\bar{\mathbf{A}} \bar{\mathbf{B}},...,\mathbf{C}\bar{\mathbf{A}}^{L_x-1} \bar{\mathbf{B}}),\\
&\mathbf{y}=\mathbf{x}*\bar{\mathbf{K}}
\end{aligned}
\end{equation}
where $L_x$ is the length of the input sequence $ \mathbf{x} $, $ \bar{\mathbf{K}} \in \textbf{R}^{L} $ is a structured convolutional kernel, and $*$ denotes a convolutional operation.

\section{A.2 Model Details}

The Mamba Encoder and Decoder can be integrated as plug-and-play modules into various collaborative perception frameworks, effectively enhancing their functionality. Additionally, they are also capable of performing single-agent object detection tasks by simply connecting the Encoder and Decoder in series. In this study, we set the number of channels \(c\) for the intermediate feature \(\mathcal{F}_i^{(t)}\) to 96 and maintained this configuration across all sub-modules. This approach significantly reduces the data transmission volume compared to models based on CNN and Transformer frameworks (\(c=192\) or more) while maintaining nearly the same level of object detection accuracy. Consequently, it improves both the computational and communication efficiency of the model.


\section{A.3 About Datasets}
We primarily conducted experiments on several datasets, including OPV2V, V2XSet.

OPV2V \cite{xuOPV2VOpenBenchmark2022a} is a large-scale simulated dataset for multi-agent V2V perception, co-simulated with Carla \cite{dosovitskiyCARLAOpenUrban2017} and OpenCDA \cite{xuOpenCDAOpenCooperative2021}. The dataset features 73 scenes across 6 road types in 9 cities, comprising 12K LiDAR point cloud frames and 230K annotated 3D bounding boxes. It includes a total of 10,914 3D annotated LiDAR frames, divided into training (6,764 frames), validation (1,981 frames), and testing (2,169 frames) sets.

V2XSet \cite{xuV2XViTVehicletoEverythingCooperative2022a} is a simulated dataset designed for V2X perception. It features 73 representative scenes with 2 to 5 connected agents and includes 11,447 3D annotated LiDAR point cloud frames, divided into training (6,694 frames), validation (1,920 frames), and testing (2,833 frames) sets.


\section{A.4 Experimental Settings}
The experiments in this paper, including tests of model computational complexity and inference speed, were conducted on an Ubuntu 22.04 server equipped with an i9 14900k CPU, two RTX 4090 GPUs, and 128GB of RAM. The models were implemented using Python 3.10 and PyTorch 2.3. Consistent with the baseline methods, the models use PointPillars (Lang et al., 2019) as the encoder. The detection range is set to \( x \in [-140.8m, +140.8m] \), \( y \in [-40m, +40m] \), with a voxelization grid size of [0.4m, 0.4m]. The hyperparameters of the network architecture are detailed in Table \ref{table-params}. If the aim is to improve model accuracy without regard to computational and communication overhead, increasing the number of feature channels would be an effective strategy. This could potentially lead to a more accurate model by leveraging the cross-agent Mamba module's ability to model long-range spatiotemporal dependencies. However, the experiments in this paper adopt a more conservative approach, balancing efficiency and accuracy to achieve an optimized trade-off.

\begin{table}[]
\centering
\caption{The network parameters of CollaMamba.}
\label{table-params}
\begin{tabular}{@{}ccc@{}}
\toprule
Model                                                                          & Parameter                                                                            & Value \\ \midrule
\multirow{14}{*}{\begin{tabular}[c]{@{}c@{}}CollaMamba\\ -Simple\end{tabular}} & epochs                                                                               & 40    \\
                                                                               & batch\_size                                                                          & 2     \\
                                                                               & lr                                                                                   & 0.002 \\
                                                                               & Mamba\_Encoder.in\_dims                                                              & 64    \\
                                                                               & Mamba\_Encoder.patch\_size                                                           & 8     \\
                                                                               & Mamba\_Encoder.stride                                                                & 4     \\
                                                                               & Mamba\_Encoder.depths                                                                & 10    \\
                                                                               & Mamba\_Encoder.dims                                                                  & 96    \\
                                                                               & Cross\_Agent\_Fusion.depths                                                          & 4     \\
                                                                               & Cross\_Agent\_Fusion.dims                                                            & 96    \\
                                                                               & Mamba\_Decoder.depths                                                                & 2     \\
                                                                               & Mamba\_Decoder.Upsample                                                              & 1     \\
                                                                               & Mamba\_Decoder.out\_conv                                                             & 1     \\
                                                                               & Mamba\_Decoder.out\_dims                                                             & 384   \\ \midrule
\multirow{7}{*}{\begin{tabular}[c]{@{}c@{}}CollaMamba\\ -ST\end{tabular}}      & epochs                                                                               & 60    \\
                                                                               & batch\_size                                                                          & 1     \\
                                                                               & lr                                                                                   & 0.001 \\
                                                                               & Trajectory\_len                                                                      & 10    \\
                                                                               & \begin{tabular}[c]{@{}c@{}}Trajectory\_Encoder.\\ depths\end{tabular}                & 8     \\
                                                                               & \begin{tabular}[c]{@{}c@{}}Feature\_Boosting.Temporal\_\\ Fusion.depths\end{tabular} & 4     \\
                                                                               & \begin{tabular}[c]{@{}c@{}}Feature\_Boosting.Spatial\_\\ Fusion.depths\end{tabular}  & 4     \\ \midrule
\multirow{7}{*}{\begin{tabular}[c]{@{}c@{}}CollaMamba\\ -Miss\end{tabular}}    & epochs                                                                               & 60    \\
                                                                               & batch\_size                                                                          & 1     \\
                                                                               & lr                                                                                   & 0.001 \\
                                                                               & Trajectory\_len                                                                      & 20    \\
                                                                               & \begin{tabular}[c]{@{}c@{}}Trajectory\_Encoder.\\ depths\end{tabular}                & 12    \\
                                                                               & \begin{tabular}[c]{@{}c@{}}Feature\_Boosting.Temporal\_\\ Fusion.depths\end{tabular} & 4     \\
                                                                               & \begin{tabular}[c]{@{}c@{}}Feature\_Boosting.Spatial\_\\ Fusion.depths\end{tabular}  & 4     \\ \bottomrule
\end{tabular}
\end{table}

\section{A.5 Training Details}
In the CollaMamba framework, the BEV encoder and detection head for the object detection task retain some 2D convolutional structures to ensure compatibility with other baseline models. Therefore, during the training of CollaMamba, we recommend pre-training the BEV encoder and detection head, then loading their pre-trained weights, focusing the training on the parameters of the Mamba Encoder, Cross-Agent Fusion, and Mamba-Decoder modules. In this study, we utilize the pre-trained weights of the BEV encoder and detection head from Where2comm-multiscale to aid in training the CollaMamba backbone network.

In the comparison experiments shown in Table \ref{table-comparison}, CollaMamba-Miss is trained under optimal communication conditions, without any miss-receiving events. To investigate robustness to poor neighbor communication, we fine-tune the trained CollaMamba-Miss model by introducing random miss-receiving intervals ranging from [1, 20] over 15 epochs. During fine-tuning, we freeze the BEV encoder, detection head, Mamba Encoder, Cross-Agent Fusion, and Mamba-Decoder, adjusting only the parameters of the cross-agent feature prediction module to help the model adapt to various miss-receiving intervals. For testing, we fix a specific miss-receiving interval, obtain a set of experimental results corresponding to a point in Figure \ref{fig-poor}, and then iterate through all possible values to generate a complete accuracy curve as the miss-receiving interval varies. This approach allows the model to achieve broad adaptability to different communication scenarios with varying miss-receiving intervals at a relatively low training cost. Further fine-tuning or training from scratch for different miss-receiving intervals could yield even better results, further improving the model's accuracy during miss-receiving events.

\section{A.6 Discussion on Model Computational Complexity}

\begin{table*}[]
\centering
\caption{Model Parameters and FLOPs of CollaMamba-ST on OPV2V.}
\label{tab:model_params_flops}
\begin{tabular}{l l l}
\toprule
\textbf{Module} & \textbf{\#Parameters or Shape} & \textbf{\#FLOPs} \\ 
\midrule
\texttt{model} & 5.349M & 0.108T \\ 
\texttt{vss\_backbone} & 3.435M & 76.932G \\ 
\texttt{vss\_backbone.encoder} & 2.33M & \textbf{27.188G} \\ 
\texttt{vss\_backbone.encoder.absolute\_pos\_embed} & (1, 50, 176, 96) & \\ 
\texttt{vss\_backbone.encoder.patch\_embed} & \textbf{0.394M} & \textbf{13.858G} \\ 
\texttt{vss\_backbone.encoder.layers} & 1.054M & 12.989G \\ 
\texttt{vss\_backbone.encoder.downsamples.0} & 37.632K & 0.341G \\ 
\texttt{vss\_backbone.decoder} & 1.105M & \textbf{49.744G} \\ 
\texttt{vss\_backbone.decoder.vss\_layers} & 0.422M & 6.089G \\ 
\texttt{vss\_backbone.decoder.upsamples.1} & 37.056K & 0.256G \\ 
\texttt{vss\_backbone.decoder.out\_layer} & \textbf{0.647M} & \textbf{43.399G} \\ 
\texttt{vss\_backbone.decoder.out\_layer.UpSample}  & 0.332M & \textbf{35.036G} \\ 
\texttt{vss\_backbone.decoder.out\_layer.Conv2d} & 0.768K & 81.101M \\ 
\texttt{fusion\_net.cross\_mamba\_blocks} & 0.477M & 0.827G \\ 
\texttt{fusion\_net.cross\_mamba\_blocks.0} & 0.119M & 0.207G \\ 
\texttt{fusion\_net.cross\_mamba\_blocks.1} & 0.119M & 0.207G \\ 
\texttt{fusion\_net.cross\_mamba\_blocks.3} & 0.119M & 0.207G \\ 
\texttt{history\_encoder} & 0.952M & 27.144G \\ 
\texttt{history\_encoder.temporal\_pos\_embedding} & (1, 1, 1, 10, 96) & \\ 
\texttt{history\_encoder.layers} & 0.659M & 24.541G \\ 
\texttt{history\_encoder.downsample\_layers.0.layer} & 37.632K & 1.706G \\ 
\texttt{history\_encoder.out\_layers} & 0.255M & 0.896G \\ 
\texttt{history\_fusion\_net.layers} & 0.477M & 1.654G \\ 
\texttt{history\_fusion\_net.layers.0} & 0.119M & 0.414G \\ 
\texttt{history\_fusion\_net.layers.1} & 0.119M & 0.414G \\ 
\texttt{history\_fusion\_net.layers.2} & 0.119M & 0.414G \\ 
\texttt{history\_fusion\_net.layers.3} & 0.119M & 0.414G \\ 
\texttt{cls\_head} & 0.77K & 81.101M \\ 
\texttt{reg\_head}  &  5.39K & 0.568G \\ 
\texttt{dir\_head} & 1.54K & 0.162G \\ 
\bottomrule
\end{tabular}
\end{table*}

Table \ref{tab:model_params_flops} provides a detailed breakdown of the parameter count and FLOPs for each sub-module within the CollaMamba-ST model, offering insights into the computational demands of different components. By using CollaMamba-ST as a representative example, the table highlights the parameter-heavy and computation-intensive components in bold black text. These components are predominantly convolutional layers, upsampling operations, and other similar processes, which are known for their higher computational complexity.

To further reduce the model's computational costs, these modules could potentially be replaced or augmented with Mamba modules that maintain linear complexity. The Mamba modules in these paper are designed to efficiently handle spatial-temporal dependencies without the exponential increase in computational requirements typically associated with traditional layers like convolutions. By transitioning more of the model's architecture to utilize Mamba's linear complexity, the overall efficiency of the model could be greatly improved, reducing both the parameter count and FLOPs without compromising performance.

 \section{A.7  Analysis of Robustness to Localization Error}

\begin{figure*}[!htb]
    \centering
    \includegraphics[width=1\textwidth]{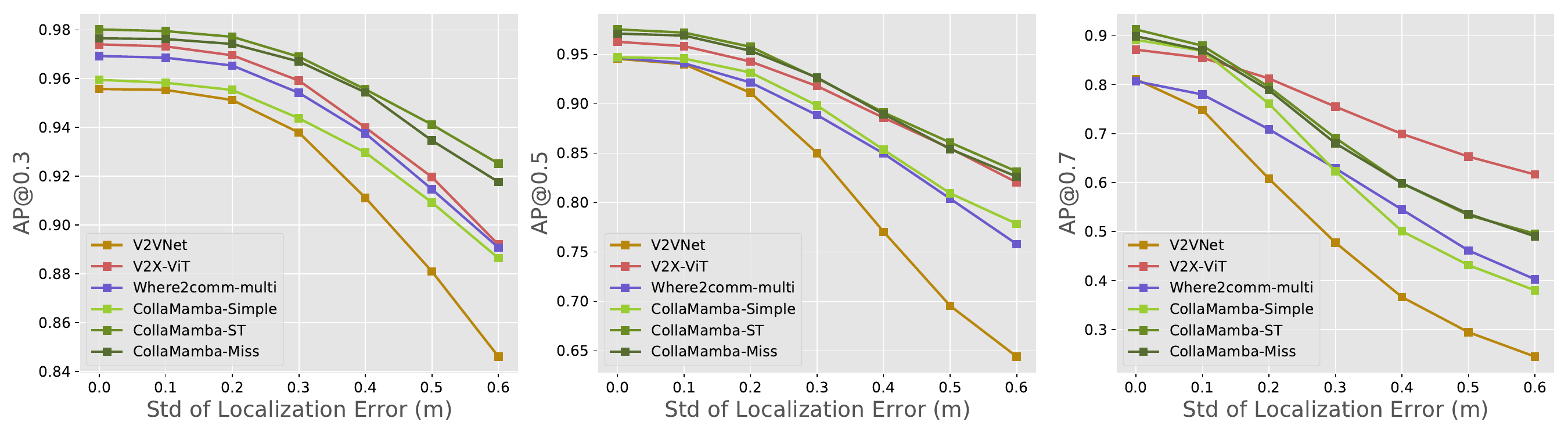}
    \caption{Robustness to localization errors on OPV2V dataset.}
    \label{fig:opv2v_noise}
\end{figure*} 

We further evaluated the performance of CollaMamba and baseline methods in scenarios with localization errors, as shown in Figure \ref{fig:opv2v_noise}. As localization errors increase, the performance of all collaborative methods decreases. However, the three CollaMamba models outperform the other compared methods to some extent, with their AP not decreasing significantly as the error range increases, demonstrating CollaMamba's robustness to localization errors. This robustness is attributed to CollaMamba's ability to capture rich long-range spatial dependencies during feature encoding and fusion, effectively modeling the semantic-level positional associations and relative spatial relationships of the target features. Even when localization errors occur, the model can still capture these correlations and positional dependencies, providing more valuable information for collaborative perception.

 \section{A.8  Qualitative Comparison Results and Visualization}

\begin{figure*}[!htb]
    \centering
    \includegraphics[width=0.99\textwidth]{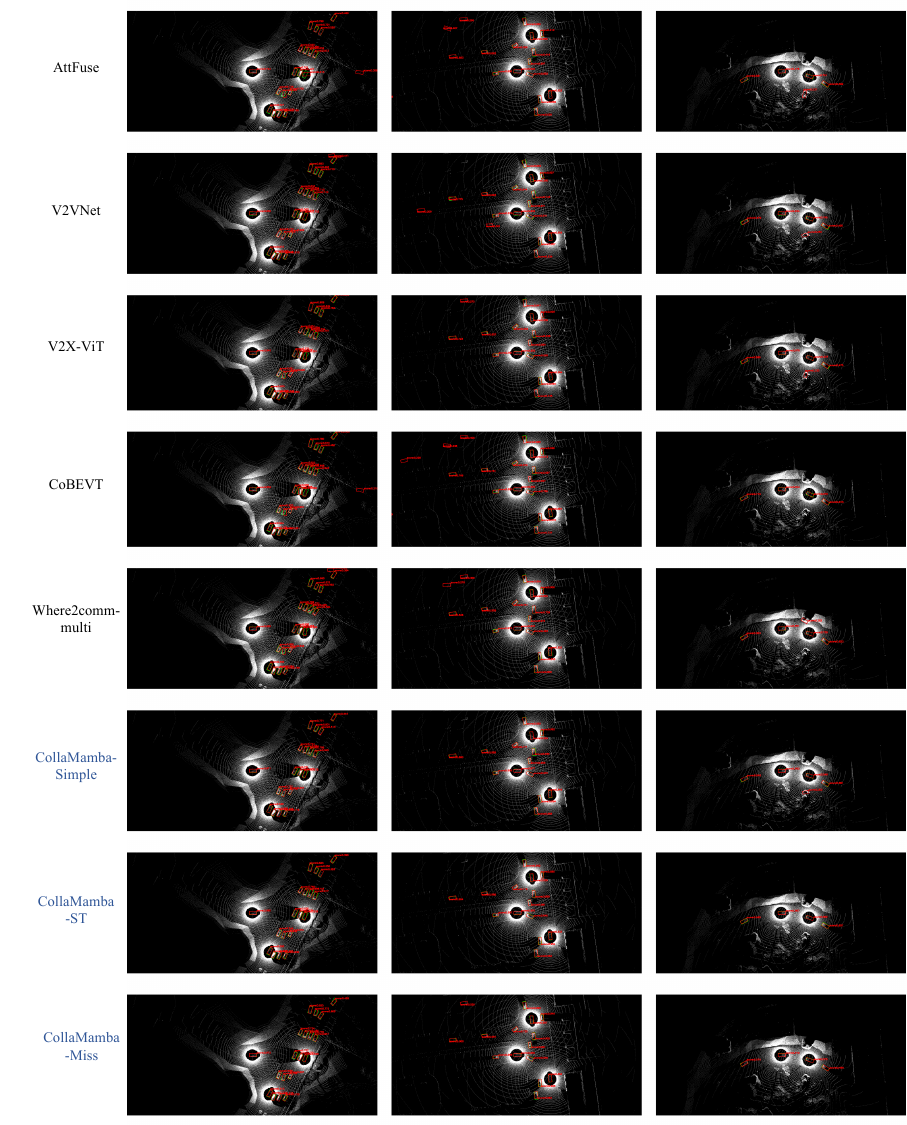}
    \caption{Visual comparison of detection results of different collaborative methods on OPV2V. The green and red boxes represent the ground-truth and the detection predictions, respectively.}
    \label{fig:vis_opv2v}
\end{figure*}

During the experiments, we randomly selected several challenging scenarios for visualization to improve qualitative comparison, as shown in Figure \ref{fig:vis_opv2v}. A close examination of the enlarged results reveals that the CollaMamba series produces more predicted bounding boxes that are well-aligned with the ground truths, while also minimizing false positives. This is particularly evident in CollaMamba-ST, which benefits from the spatial-temporal historical trajectory encoder and feature boosting module. These components enable the model to refine and enhance uncertain or ambiguous features in the current frame by leveraging contextual cues from long-term historical spatial-temporal sequences, leading to more accurate results.

\end{document}